\documentclass[conference]{IEEEtran}
\IEEEoverridecommandlockouts
\usepackage{cite}
\usepackage{amsmath,amssymb,amsfonts}
\usepackage{graphicx}
\usepackage{mathtools}
\usepackage{textcomp}
\usepackage{xcolor}
\usepackage{url}
\usepackage{amsthm}
\usepackage{times}
\usepackage{multirow}
\usepackage{epsfig}
\usepackage{graphicx}
\usepackage{amsmath}
\usepackage{amssymb}
\usepackage{booktabs}
\usepackage{algpseudocode}
\usepackage{algorithm}
\usepackage{graphicx}
\usepackage[affil-it]{authblk}

\long\def\invis#1{}

\def\BibTeX{{\rm B\kern-.05em{\sc i\kern-.025em b}\kern-.08em
    T\kern-.1667em\lower.7ex\hbox{E}\kern-.125emX}}
\begin{document}

\title{\textit{ChatSim}: Underwater Simulation with Natural Language Prompting\\
\thanks{Corresponding author: Xiaomin Lin: xlin01@umd.edu}
}

\author{Aadi Palnitkar, Rashmi Kapu, Xiaomin Lin, Cheng Liu, Nare Karapetyan, Yiannis Aloimonos}
\affil{Maryland Robotics Center, University of Maryland}
\maketitle
\thispagestyle{empty}
\pagestyle{empty}

\begin{abstract}
Robots are becoming an essential part of many operations including marine exploration or environmental monitoring. However, the underwater environment presents many challenges, including high pressure, limited visibility, and harsh conditions that can damage equipment. Real-world experimentation can be expensive and difficult to execute. Therefore, it is essential to simulate the performance of underwater robots in comparable environments to ensure their optimal functionality within practical real-world contexts. \textit{OysterSim} generates photo-realistic images and segmentation masks of objects in marine environments, providing valuable training data for underwater computer vision applications. By integrating ChatGPT into underwater simulations, users can convey their thoughts effortlessly and intuitively create desired underwater environments without intricate coding. \invis{Moreover, researchers can realize substantial time and cost savings by evaluating their algorithms across diverse underwater conditions in the simulation.} The objective of \textit{ChatSim} is to integrate Large Language Models (LLM) with a simulation environment~(\textit{OysterSim}), enabling direct control of the simulated environment via natural language input. This advancement can greatly enhance the capabilities of underwater simulation, with far-reaching benefits for marine exploration and broader scientific research endeavors.
\end{abstract}
\section{Introduction}
\label{sec:intro}

\invis{Underwater robotics has been one of the most dynamic fields of research in the recent past. It gives scope for exploration and research in the depths of the oceans, which make up more than two-thirds of the earth's surface and have been largely unexplored.} With the advancements in the field of technology and Artificial Intelligence (AI), various Unmanned Underwater Vehicles (UAVs) and Remotely Operated Vehicles (ROVs) have been developed to study and monitor underwater environments. The underwater robots have are being used in various applications, including ocean exploration, marine science, underwater archaeology\cite{karapetyan2021human}, environmental monitoring\cite{dunbabin2012robots, karapetyan2018multi}, and underwater infrastructure inspection and maintenance\cite{la2017development}. As technology advances and research progresses, underwater robots are becoming increasingly capable, versatile, and valuable tools for understanding and harnessing the potential of the underwater world.

However, despite its significance, the development of solutions for underwater robots and deployment of such systems remains  challenging due to the high water pressure, harsh tidal conditions, and limited visibility. In designing and deploying algorithmic solutions in robotics realistic physics-based simulations play a vital role. For underwater environments, OysterSim~\cite{lin2022oystersim} presents a simulation platform that can be used to create photo-realistic image datasets with multiple sensor data and the ground truth location of a Remotely Operated Vehicle (ROV). This data can be used to develop and test algorithms to study and detect oysters. The motive of this paper is to integrate Large Language Models'(LLMs) interactive communication capabilities with the BlueROV, a popular AUV in this simulation environment, aiming to enable human operators to easily create simulated underwater environments simply using natural language, without the need for writing a code.

\begin{figure}
 \centering
{\includegraphics[width=0.4\textwidth]{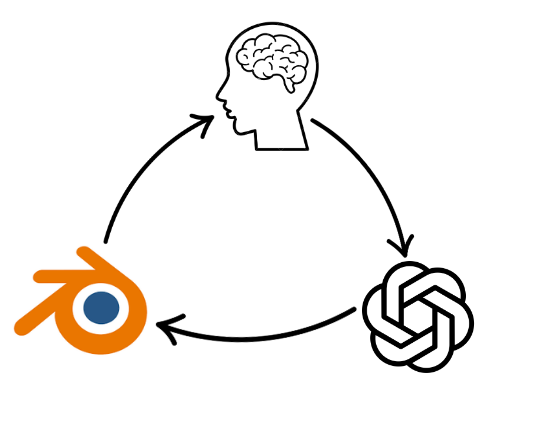}}
\caption{\label{fig:interaction} Interaction between the user, Natural Language Model - ChatGPT, and the simulation environment - Blender. }
\label{fig:1}
\end{figure}

\begin{figure*}[ht!]
\includegraphics[width=\textwidth]{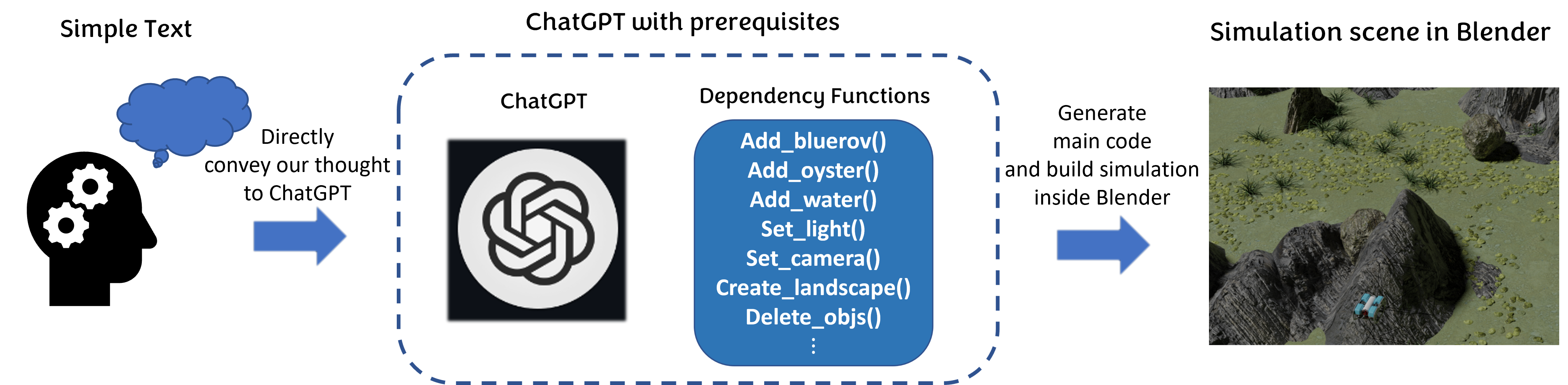}
\centering
\vspace{-3mm}
\caption{ Pipeline to create a simulation scene.}
\label{fig:pipeline}
\vspace{-3mm}
\end{figure*}

In 2022, Brown et al.~\cite{brown2020language} introduced a versatile concept with extensive potential for robotics applications that facilitate interactive communication between LLMs and humans. This interaction, depicted in Fig~\ref{fig:interaction}, enables the incorporation of objects into simulation scenes and the execution of specific commands. Following the same concept, Vemprala et al. ~\cite{vemprala2023chatgpt} integrates ChatGPT and a robotics simulator, enabling robots to perform complex tasks using natural language commands. Similarly, the Large Language Model can have significant benefits for marine robotics and beyond. In this work we present \textit{ChatSim} platform (Fig~\ref{fig:pipeline}), that integrates LLMs into an underwater simulation environment, enabling users to create scenes with different properties or direct the robot's actions based on specific natural language input.  In contrast to conventional methods, ChatSim demands less time and coding expertise for the creation or modification of an underwater scene. The proposed \textit{ChatSim} framework combines ChatGPT LLM and Blender-based\texttrademark ~\cite{blender2018blender} underwater OysterSim simulation. Within the \textit{OysterSim}, the simulation incorporates various underwater elements such as oysters, rocks, grass, coral, shipwrecks, and other objects, nestled on the seabed. An agent, BlueROV, is also integrated into the environment which moves around according to the user commands and captures photos at regular intervals. These photos can be further used for testing algorithms for oyster detection and other marine applications.  We open-source\footnote{\url{https://github.com/apalkk/PRG-Underwater-Simulation}} our \textbf{\textit{ChatSim}} framework and data associated with this work to accelerate further research.

 The rest of this paper is structured as follows. We will first give an overview of the literature relevant to this work in Section \ref{section:related_work}. Then Section \ref{section:problem_formulation} presents details of our proposed approach and its components. In Section \ref{section:Experiments_and_results} we demonstrate several applications and experiments. Finally, the concluding thoughts are drawn in Section \ref{section:conclusion} with future work remarks. 

\section{Related Work}
\label{section:related_work}

This section begins by examining simulations and datasets relevant to underwater environments. Subsequently, we delve into the realm of Transformers and LLMs and then explore how leveraging LLMs within simulations can significantly enhance the overall user experience.

\subsection{Simulations and Datasets}
AirSim~\cite{airsim}, a simulator built on Unreal Engine, aims to propose a solution for the expensive and time-consuming process of developing and testing algorithms for autonomous vehicles in the real world.  Additionally, recent advances in machine intelligence and deep learning require a large amount of annotated training data in a variety of conditions and environments. The paper introduces a novel simulator developed on Unreal Engine, providing realistic simulations for both physical and visual aspects. It features a high-frequency physics engine for real-time hardware-in-the-loop (HITL) simulations, supporting widely used protocols like MavLink. The simulator is designed from the ground up to be extensible to accommodate new types of vehicles, hardware platforms, and software protocols.
As far as underwater robotics is concerned, Zwilgmeyer et al.~\cite{zwilgmeyer2021creating} use Blender~\cite{blender2018blender} to create underwater data containing ground truth and sensor data. They developed a framework to create and track sensor trajectories by employing a physical model and control system, resulting in diverse underwater scenes with a wide range of features and complexities, encompassing everything from simple sandy bottoms to intricate geometries with occlusion.
UUV Simulator~\cite{2016uuv}, a Gazebo-based~\cite{gazebo} package, creates a simulation environment for underwater intervention and multi-robot simulation. It provides a detailed description of the software structure, simulation modules, and applications used to interact with the simulation environment via Robot Operating System~\cite{ros}.
Another such work is of HoloOcean~\cite{2022holoocean}, which is an open-source underwater simulator built on Holodeck~\cite{HolodeckPCCL} using Unreal Engine 4. It is designed to facilitate the testing and development of algorithms for autonomous underwater vehicles (AUVs). The simulator includes features such as multi-agent support, various sensor implementations, and simulated communications support.
OysterSim ~\cite{lin2022oystersim} is a simulated environment that can be used to create photo-realistic image datasets with multiple sensor data and ground truth locations of a remotely operated vehicle (ROV). It provides a new benchmark suite for the underwater community by offering a simulated environment that can be used to develop and test navigation and path-planning algorithms with oyster-based localization. The images generated from variants of OysterSim can be used to detect BlueROV \cite{lin2023seadronesim}, oysters\cite{lin2023oysternet}, and whales\cite{gaur023whale}.
Nonetheless, customizing these simulations to align with the specific needs of users demands a substantial level of domain expertise and would entail a significant time investment in programming. Hence, we use the simulation environment of the OysterSim in Blender and propose to extend its functionality by enabling users even without prior knowledge of the Blender engine and programming skills, to interact with the environment. we will spotlight the utilization of transformers and LLMs in robotics in the following paragraphs. 

\subsection{Transformers and LLMs in Robotics}
The Transformer architecture proposed in~\cite{attention} has played a pivotal role in shaping the NLP (Natural Language Processing) landscape and advancing the field of AI. It has laid the foundation for many of the most successful and influential language models used today. LLMs overcome the challenges posed by classical symbolic AI because they have the ability to understand the context and generate new pieces of code based on that context. Incorporating this feature into simulation environments offers significant advantages as it expands the potential user base substantially.
The integration of LLMs has proven to be immensely beneficial across a wide array of fields, fostering significant advancements and breakthroughs in diverse research applications. Its versatile capabilities have been instrumental in driving progress and innovation across various domains. The following section highlights diverse LLM applications across various domains.

Microsoft Autonomous Systems and Robotics Research released the AirSim simulator with ChatGPT integration~\cite{vemprala2023chatgpt}, called 'PromptCraft', so that the users can easily convey their thoughts, for example, move a drone as desired, in the simulation by prompting the ChatGPT, which is very similar to our application.  GesGPT~\cite{2023gesgpt} proposes a novel approach to gesture generation centered around text parsing using LLM, which utilizes the robust semantic analysis capabilities of Large Language Models like GPT-3 to produce semantically rich co-speech gestures. The paper discusses the potential of introducing LLMs into the field of embodied intelligence and gesture synthesis. Wake et al.~\cite{wake2023chatgpt} introduced a novel method for translating natural-language instructions into executable robot actions using OpenAI's ChatGPT. It allows for customizable input prompts, multi-step task plans, and conversational feedback and can be used in various environments, including domestic settings, to control robots and perform tasks based on natural language instructions. We propose the utilization of LLM to interpret user instructions in natural language and generate corresponding Python scripts in response. This script would utilize the provided functions, and we present a comprehensive demonstration of this process in sections \ref{section:problem_formulation} and \ref{section:Experiments_and_results}.

\section{Approach}
\label{section:problem_formulation}

The primary objective of \textit{ChatSim} is to enhance the user's interaction with the simulation, democratizing access to all users and enabling them to create customized underwater simulations and execute robotic actions without requiring knowledge of Blender or a programming language. 
\textit{ChatSim} enables users to seamlessly incorporate various types of objects into the simulation scene and make necessary scene modifications according to their specific requirements. Furthermore, the robot/agent can navigate the environment and execute designated actions based on natural language inputs provided by users. Through the fusion of these technologies, we can create a comprehensive solution for developing and testing algorithms that facilitate the efficient operation of robots within the underwater domain.
 \begin{figure}
 \centering
{\includegraphics[width=0.5\textwidth]{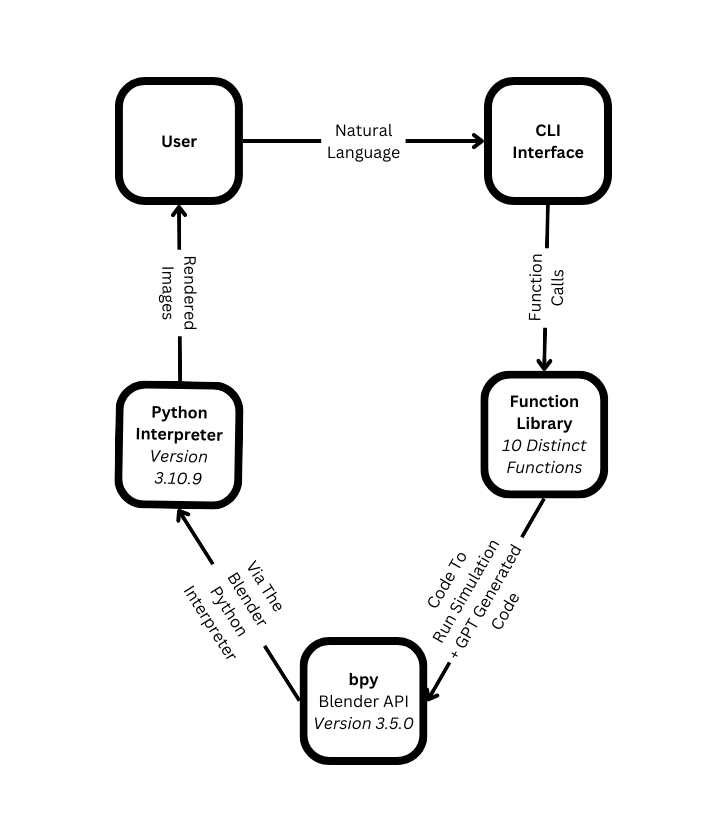}}
\caption{\label{fig:flowchart1} The flowchart of the actions taking place during the execution. The system eliminates the engineer 'middleman' by integrating LLMs into the pipeline. }
\label{fig:4}
\end{figure}

To begin with, we first present one environment example, go through some implementation details and highlight specific functionalities. Subsequently, we will elaborate on the execution pipeline. Finally, we conclude this section by summarizing potential drawbacks and possible future directions.

\begin{figure*}[ht!]
\includegraphics[width=0.95\textwidth]{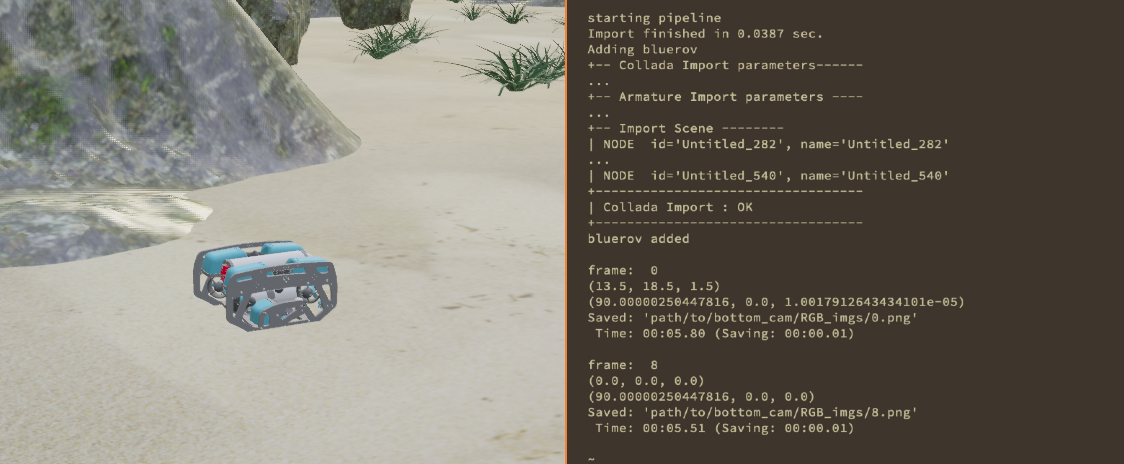}
\centering
\vspace{-1mm}
\caption{ An image depicting the simulation's initial startup process. To the left is the initial position and orientation of the agent (BlueROV) and to the right is the image depicting how the simulation starts on a terminal window.}
\label{fig:prompt}
\vspace{-4mm}
\end{figure*}

\subsection{Implementation Details}

We focus on allowing \textit{ChatSim} to automatically test and develop algorithms that enable robots to operate effectively in the underwater environment. We created a function library where some sample functions can be seen in Table~\ref{tab:functions}, giving LLM the freedom to control the simulation while restraining it to only use our functions. The functions are explicitly defined in the prompt, and clear and concise instructions are provided to prevent it from creating its own functions. 

\begin{table}[hb!]
\centering
\vspace{-1mm}
\caption{List of Prompt Functions}
\begin{tabular}{|p{3.2cm}|p{5cm}|}
\hline
\textbf{Function} & \textbf{Description} \\
\hline
set\_bot\_position(points) & Takes a tuple as input indicating the X, Y, and Z coordinates the user wants the bot to move to. \\
\hline
get\_position(object\_name) & Takes a string as input indicating the name of an object of interest, and returns a vector of 3 floats indicating its X, Y, Z coordinates. \\
\hline
get\_bot\_position() & Gets the current X, Y, Z coordinates of the drone. \\
\hline
set\_yaw(angle) & Set the yaw angle of the drone (in degrees). \\
\hline
set\_pitch(angle) & Sets the pitch angle of the drone (in degrees). \\
\hline
set\_roll(angle) & Sets the roll angle of the drone (in degrees). \\
\hline
put\_object(name,(x,y,z),(yaw, pitch, roll)) & Adds objects to the simulation by taking in the name of the object, coordinates, and orientation angles. \\
\hline
delete\_objects\_in\_range(...) & Deletes mesh objects within a specified coordinate range. \\
\hline
put\_bot\_switch(coordinates) & Adds a non-agent BlueROV to the specified coordinates. \\
\hline
\end{tabular}
\label{tab:functions}
\end{table}

As an illustrative example, we showcase the simulation's capability by designing the terrain to emulate a shallow oyster reef environment. Oysters and rocks are randomly incorporated into the simulated seabed. The simulation saves the photos taken from cameras mounted on the front of BlueROV at specified time-frame intervals within the \textit{ChatSim}. One of the benefits of this approach is that it facilitates the extension of the simulation capabilities by adding new functions to the library. An additional advantage of employing a function library is its ability to restrict the extent of the LLM's impact on the simulation, thereby mitigating the potential for errors arising from the generated code, although implementing such functionality in the simulation would be straightforward. The approach we have adopted for implementing our system ensures the absence of a feedback loop, preventing any exposure of simulation data to LLM, which is elaborated further, in the later part of this section.
Before every natural language interaction, we use a system prompt to instruct the LLM on what kind of responses we want and our function library. System prompts are special messages used to control the behavior of LLMs and define the bounds and style of their responses. We explicitly use well-defined statements to allow it to understand how every response must be structured and what code it is allowed to respond with. We also have safeguards in our code to prevent any unintended behavior. An example of this is a regex created to extract code from the GPT-generated text markdown in case it responds with text as well as code. 

The main functionalities of proposed ChatSim can be sumarised as follows:

\begin{itemize}
    \item \textbf{Change agent}: By providing necessary function descriptions to the simulation interface and 3D models of the agents, \textit{ChatSim} will be able to import BlueROV or any other agents to perform specific tasks. Here in this case, the BlueROV model is imported into the simulation with control over all six degrees of freedom, ensuring comprehensive exploration and manipulation possibilities. 

\item \textbf{Change Scene}: Moreover, the objects in the background are separable, allowing for easy switching of the background scenes and exploring the potential of simulating various underwater research scenarios. Additionally, the simulation includes various other elements such as coral, grass, oysters, shipwrecks, and more, using pre-existing 3D models and textures which would allow the user to simulate other underwater environments. Furthermore, the option to delete objects within a specified range offers a convenient way to modify the environment as needed. Lastly, users also have the capability to modify the water's properties, including parameters such as watercolor and turbidity.

\end{itemize}

\invis{Following the implementation details, we will proceed to showcase the execution pipeline within the \textit{ChatSim} environment. This demonstration aims to provide users with a clear and intuitive understanding, allowing them to effortlessly engage with the simulation.}
\subsection{Execution Pipeline} 
The execution pipeline consists of the following steps: 
First, the simulation starts with importing all the necessary libraries, functions, and models for elements in the simulation. Sand, rocks, hills, oysters, corals, shipwrecks, and other elements along with the agent (robot) are added to the underwater scene. The agent moves around and takes pictures from the cameras mounted on it. If input mode is specified in the settings, a prompt is fed to LLM, enabling real-time user input as the simulation progresses, as shown in the right-hand side of Fig~\ref{fig:prompt}. This requires a unique API key that can be generated on the OpenAI~\cite{openai2023} platform. Then the prompt will appear and instructions can be given in Natural Language for which the LLM generates an equivalent Python script.
This generated Python script is then executed and accordingly reflected in the simulation environment. Photos from the cameras on BlueROV are taken and saved. A detailed demonstration of the same is shown in section \ref{section:Experiments_and_results}. 
\begin{figure*}[ht!]
\includegraphics[width=\textwidth]{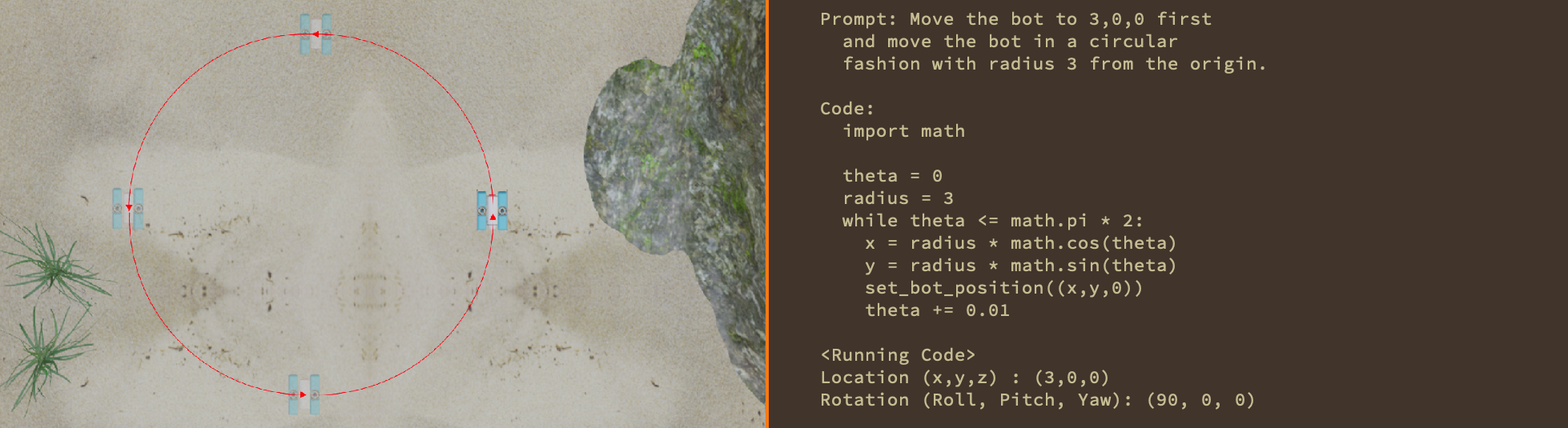}
\centering
\vspace{-3mm}
\caption{ The image on the left depicts the trajectory taken by the BlueROV when given a command to follow a circular path with a radius of 3 units, as illustrated in the image on the right.}
\label{fig:move2}
\vspace{-5mm}
\end{figure*}
As we mentioned in the previous paragraph - the two different modes for \textit{ChatSim} are:
\begin{itemize}
    \item \textbf{With input mode}: where the simulation stops after a predefined interval and accepts natural language input from the terminal which is then sent to the LLM which returns code output. This is demonstrated in the flowchart Fig~\ref{fig:flowchart1}.
    
\item \textbf{Without input mode} allowing only predefined instructions (given in the settings file) to be executed in the same intervals where the simulation pauses. It is possible to give predefined instructions while the input mode is on, but the predefined instructions will execute before accepting input.
\end{itemize}
\invis{After discussing the implementation details and the execution pipelines, we will conclude this section by summarizing potential risks and considerations.}

\subsection{Potential Risks}
While acknowledging the benefits of various LLM tools, it is important to address the potential risks associated with their usage. We have implemented closed perception-action loops in order to prevent any leakage of data from the simulation and the function library and our explicit instructions in the system prompt restrict ChatGPT's actions to our library's predefined function calls. This ensures that arbitrary code execution is not allowed, thus minimizing security vulnerabilities.

It must be emphasized that caution should be exercised when directly implementing code generated by any LLM, due to its susceptibility to errors. However, we anticipate future developments in closed-loop systems that will enable users to mitigate the risk of erroneous code while maintaining precise control over simulations without relying on custom-written code. As researchers continue exploring this field and extending the functionality within our toolkit, it is recommended they adopt a similar methodology for achieving successful outcomes and efficient utilization.

\section{Experiments}
\label{section:Experiments_and_results}
In this section, we present multiple experiments that demonstrate how the LLM generates Python code according to the user instruction, and the consequential effects it has on the Blender simulation. Fig \ref{fig:prompt} shows the terminal interface displayed before the user can give the instructions. The program creates a simulation and adds objects underwater and saves the images from the simulation, as seen in the figure. After following the execution pipeline, the user can give natural language instructions.

\invis{The simulation's frame limit is set at 1000 frames, allowing around 125 actions (at an 8-frame interval). This can be adjusted by modifying frame settings. In the case of Fig~\ref{fig:move2}, the given theta generates approximately 62 actions, requiring 496 frames at an 8-frame interval.
\subsubsection{Experiment 1}To explore the surroundings from the BlueROV in a particular position, we need to be able to rotate it in its position and take pictures at regular intervals from the cameras mounted on the ROV. In Fig \ref{fig:prompt}, the position and orientation of the BlueROV before prompting the ChatGPT to rotate it is shown. Subsequently, after instructing to set the yaw of the ROV to 100 degrees, the position and the terminal output are shown in Fig \ref{fig:rotation_fin}. We can see that the Python script is generated to call the function 'set\_yaw()' and the pictures from the cameras mounted on the BlueROV are being saved.}
\subsubsection{Experiment 1} 
\begin{figure*}[ht!]
\includegraphics[width=\textwidth]{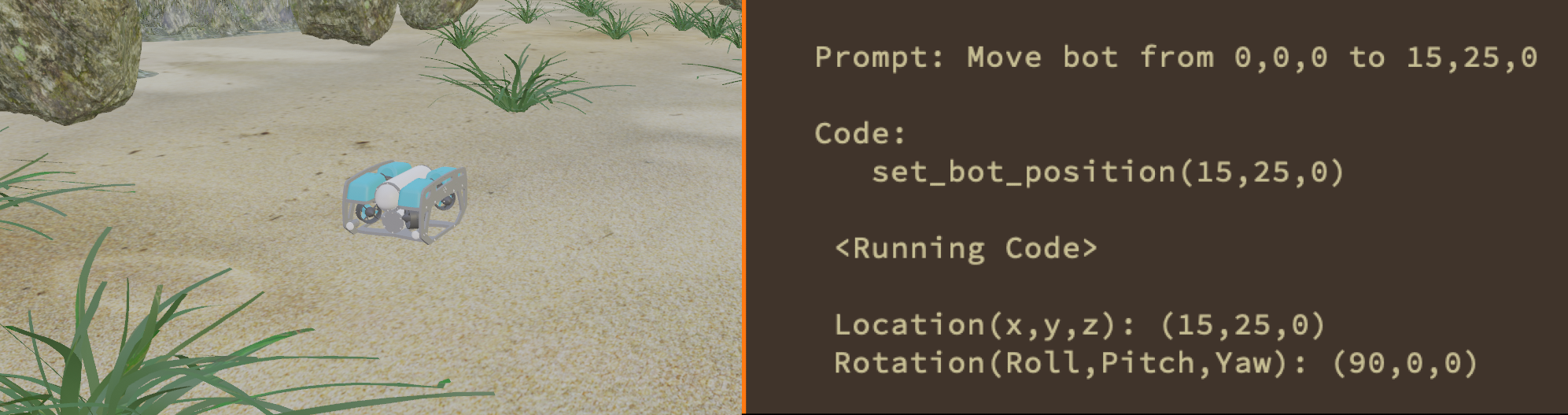}
\caption{ The location of BlueROV is changed after prompting LLM to move it from 0,0,0 to 15,25,0, as seen on the terminal.}
\label{fig:move_fin}
\end{figure*}

In order to explore a particular trail in the environment, we defined a functions to change the robot's position in the scene - location and the heading. The initial position of the bot can be seen in \ref{fig:prompt}. When the user prompts the GPT to move the BlueROV from initial coordinates to 15,25,0, we can see the change in the final position in Fig \ref{fig:move_fin}, for which 'set\_bot\_position()' function is called, and the final coordinates are passed as arguments. During this, photos from cameras mounted on the ROV are taken at regular intervals and saved.
\subsubsection{Experiment 2} 
\begin{figure*}[h]
\includegraphics[width=\textwidth]{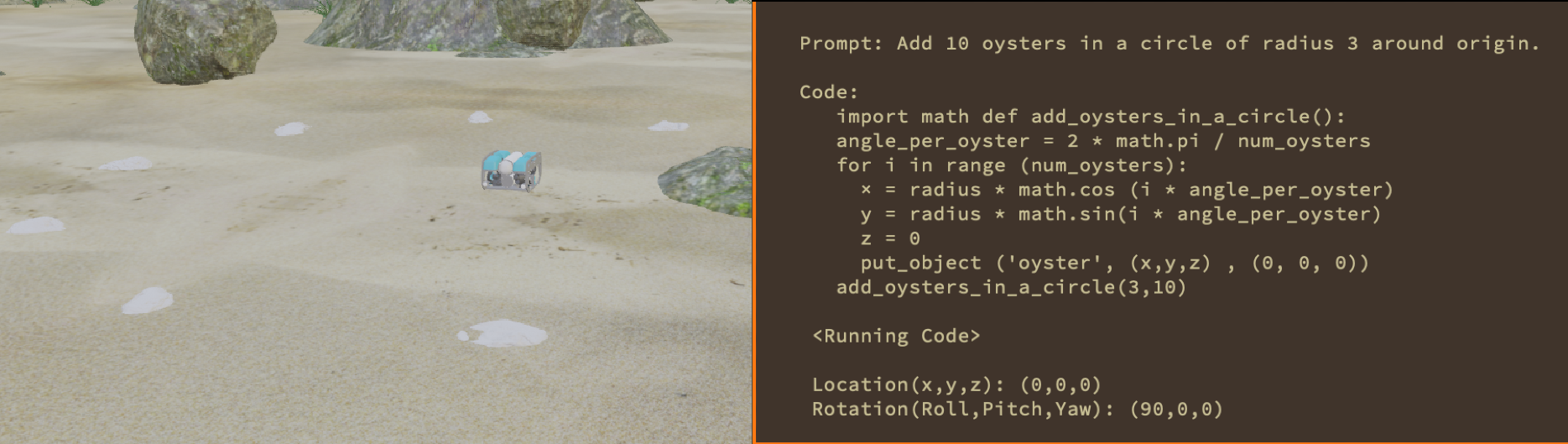}
\centering
\caption{ Scene as oysters are added as a circle around the BlueROV.}
\label{fig:addition_final}
\end{figure*}

We defined another function called 'put\_object()', which facilitates the user to add more objects in the simulation. As seen in Fig \ref{fig:addition_final}, the user wishes to add ten oysters on a circle of radius 3 around the origin. It can be seen in the terminal window that the function called by ChatGPT to do so is 'put\_object()' with the name of the element, position, and orientation of the oysters to be added. This function is called ten times to add ten oysters.
\subsubsection{Experiment 3} 
\begin{figure*}[h]
\includegraphics[width=\textwidth]{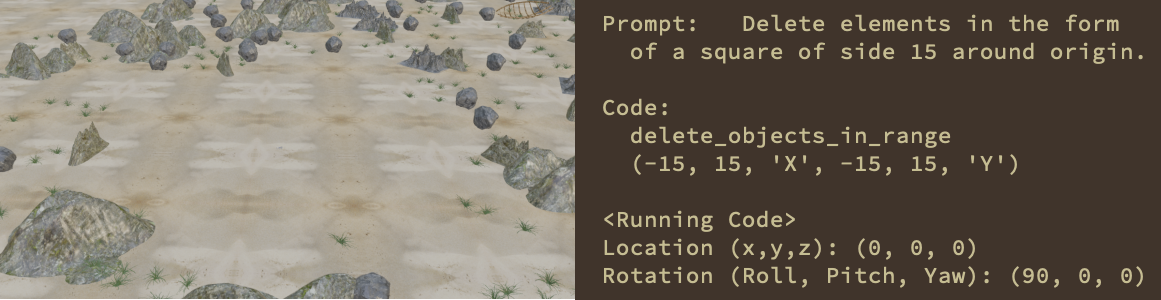}
\centering
\caption{ Simulation scene after prompting to delete all objects that lie on or within the square of side 15 with origin as the center.}
\label{fig:deletion_final}
\end{figure*}

The user can also delete elements by specifying the area as seen in Fig~ \ref{fig:deletion_final}, where a portion (square of side 15 with origin as the center)can be seen clearly in the simulation. As seen on the terminal window to the right, the function called for this is 'delete\_objects\_in\_range()', with the lower and upper bound of the range passed as arguments.
\subsubsection{Experiment 4} Fig~\ref{fig:move2}, illustrates an experiment showcasing the robot's circular trajectory generated through the LLM. This method aids in systematically exploring an area by moving along its perimeter and capturing images at regular intervals. The angular parameter 'theta' is utilized, incriminating in 0.01 intervals until it reaches 360 degrees, effectively calculating the positions for each iteration relative to the origin. The simulation's multi-action movement is achieved by queuing and executing multiple function calls successively. 

Numerous additional experiments can be conducted within the \textit{ChatSim} framework. Users have the flexibility to fine-tune all object and water parameters, enabling a wide range of exploratory scenarios. The environment remains open for user exploration, with potential adjustments to functions and prompts to align with specific requirements. We are readily available to offer assistance in tailoring the system to users' preferences.

\invis{\begin{figure*}[ht!]
\includegraphics[width=\textwidth]{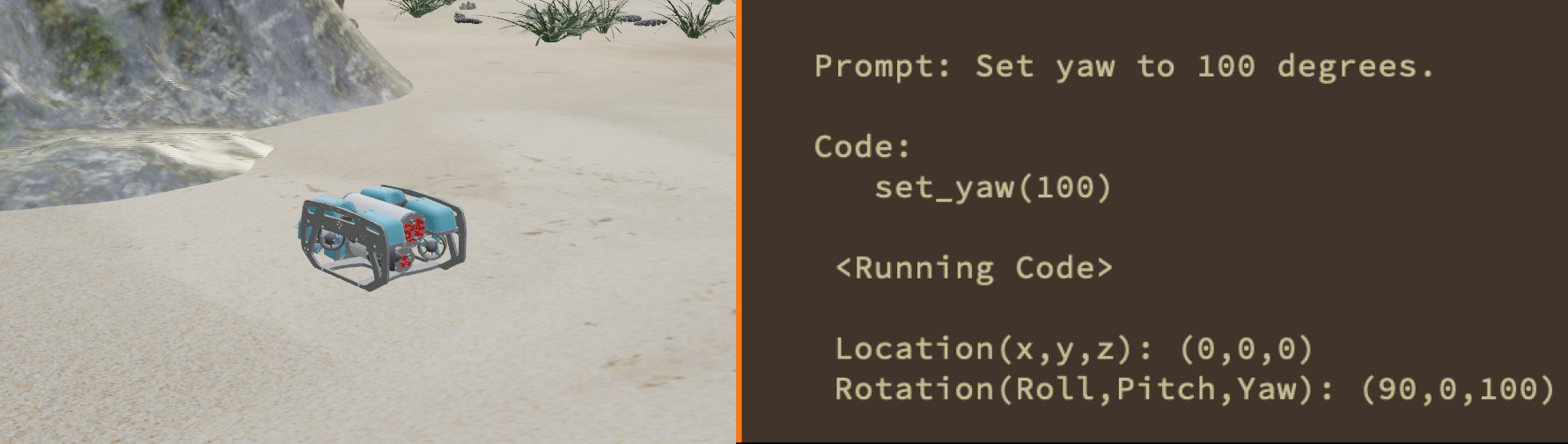}
\centering
\caption{ BlueROV is rotated after prompting LLM to change the yaw.}
\label{fig:rotation_fin}
\end{figure*}}
\section{Conclusions and Future Work}
\label{section:conclusion}
In this work, we present an integrated framework that combines LLM, with an underwater simulation to extend its applications. \textit{ChatSim} introduces a no-code simulation approach that allows individuals with basic math skills to easily comprehend and utilize the simulation. This integration is supported by the use of function libraries and design principles that provide the necessary context for this innovative approach. One of the key features of our framework is the ability for users to interact with the simulation using natural language. By leveraging the power of LLM, users can input commands and queries in natural language, eliminating the need for complex programming languages or technical expertise. The simulation can be viewed in real-time by the images generated within Blender. These images, subject to human validation, not only enhance the simulation's utility and accessibility but also contribute to advancing marine research.

To enhance the realism and accuracy of the simulations, we could integrate more real-world data into the framework. This could involve sourcing and incorporating relevant data sets from marine research initiatives, environmental monitoring, and other sources. The simulation could then provide users with the opportunity to explore and analyze scenarios based on actual data. More sensors can be introduced for the simulation. For instance, multi-beam sonar has been widely used for marine tasks. We are coming up with a solution for that. 

\section*{ACKNOWLEDGMENT}
This work is supported by "Transforming Shellfish Farming with Smart Technology and Management Practices for Sustainable Production" grant no. 2020-68012-31805/project accession no. 1023149 from the USDA National Institute of Food and Agriculture.
\vspace{3mm}
\bibliographystyle{IEEEtran}
\bibliography{refs}
\end{document}